\documentclass{article}

\PassOptionsToPackage{numbers, compress}{natbib}
\usepackage[final]{neurips_2024_ml4ps}

\usepackage{amsmath}

\usepackage{todonotes}
\usepackage[utf8]{inputenc} % allow utf-8 input
\usepackage[T1]{fontenc}    % use 8-bit T1 fonts
\usepackage{hyperref}       % hyperlinks
\usepackage{url}            % simple URL typesetting
\usepackage{booktabs}       % professional-quality tables
\usepackage{amsfonts}       % blackboard math symbols
\usepackage{nicefrac}       % compact symbols for 1/2, etc.
\usepackage{microtype}      % microtypography
\usepackage{xcolor}         % colors
\usepackage{listings}
\usepackage[inline]{enumitem}

\title{Testing Uncertainty of Large Language Models for Physics Knowledge and Reasoning}

\author{%
Elizaveta Reganova \\
  Helmholtz AI Team Matter\\
  Helmholtz-Zentrum Dresden-Rossendorf\\
  01328 Dresden Germany\\
  \texttt{lisa.reganova@gmail.com} \\
  \And
Peter Steinbach \\
  Helmholtz AI Team Matter\\
  Helmholtz-Zentrum Dresden-Rossendorf \\
  01328 Dresden Germany\\
  \texttt{p.steinbach@hzdr.de} \\
}

\begin{document}

\maketitle

\begin{abstract}
Large Language Models (LLMs) have gained significant popularity in recent years for their ability to answer questions in various fields. However, these models have a tendency to ''hallucinate'' their responses, making it challenging to evaluate their performance. A major challenge is determining how to assess the certainty of a model's predictions and how it correlates with accuracy. In this work, we introduce an analysis\footnote{All code and data is made available. For details, see Appendix \ref{app:sec:code}.} for evaluating the performance of popular open-source LLMs, as well as gpt-3.5 Turbo, on multiple choice physics questionnaires. We focus on the relationship between answer accuracy and variability in topics related to physics. Our findings suggest that most models provide accurate replies in cases where they are certain, but this is by far not a general behavior. The relationship between accuracy and uncertainty exposes a broad horizontal bell-shaped distribution. We report how the asymmetry between accuracy and uncertainty intensifies as the questions demand more logical reasoning of the LLM agent, while the same relationship remains sharp for knowledge retrieval tasks.
\end{abstract}

\section{Introduction}

Large language models (LLMs) have demonstrated remarkable performance across various text generation tasks, including question answering \citep{openai2024gpt4technicalreport, kasneci2023chatgpt, zhuang2023toolqadatasetllmquestion}. However, despite their impressive power and complexity, the capabilities of LLMs are inherently limited. These limitations stem from the finite nature of their training data, as well as the models' intrinsic memorization and limited reasoning capacities. Reliability is a critical component of LLM trustworthiness \citep{liu2024trustworthyllmssurveyguideline}. To build user trust, it is essential that models provide clear and accurate answers, preventing the spread of misinformation. One of the most significant challenges facing LLMs is the tendency to generate hallucinated responses \citep{farquhar2024detecting}. In tasks like question answering, it is crucial to determine when we can trust the outputs of these models. Despite the recent advancements in natural language generation, there remains limited understanding of uncertainty in foundation models \citep{lmpolygraph}.

Uncertainty estimation involves quantifying the degree of confidence in the predictions made by a machine learning model \citep{H_llermeier_2021, gawlikowski_survey_2022, abdar_review_2021}. Without proper measures of uncertainty, it is difficult to rely on generated text as a trustworthy source of information. A common approach to evaluating model performance is through the use of question-answering (QA) benchmarks, which come in various formats \citep{li2024multiplechoicequestionsreallyuseful}. Among available formats, multiple-choice questions, which present multiple candidate answers alongside the input question, are the most popular. They offer a straightforward and efficient means of assessing model performance \citep{li2024multiplechoicequestionsreallyuseful}.
%These include True/False questions (TFQs), where models predict the correctness of a given statement; multiple-choice questions (MCQs), which present multiple candidate answers alongside the input question; and long-form generation questions (LFGQs), where the generated response may span multiple sentences. A

Uncertainty estimation for machine learning has progressed to a well-studied field, particularly in the context of classification and regression tasks \citep{gawlikowski_survey_2022, gal2016uncertainty}. In this work, we make an effort to assess the uncertainty of answers by an LLM agent on a physics specific multiple choice question and answer dataset \citep{mlphys101}. Our contributions are as follows:
\begin{enumerate*}[label={(\alph*)}]
    \item we obtain answers to high-school grade physics questions by 3 open-source and one closed-source LLM,
    \item we compare the variation of answers between abovementioned LLMs and
    \item we analyse the accuracy-certainty trade-off for each LLM in five question categories.
\end{enumerate*}
    
To our knowledge, this is the first publication which focuses on the trustworthiness of LLM answers in physics reasoning and physics related knowledge retrieval.
%
%In this work, we simplify the model's output by framing it as a classification task, allowing us to estimate uncertainty by evaluating the variation among regenerated responses to the same question.

\section{Method}
\label{sec:method}

\subsection{Dataset}
\label{ssec:dataset}

The \texttt{mlphys101} dataset \citep{mlphys101} for our study consists of $823$ university-level physics multiple-choice questions in English, each with five possible answers, among which one is correct. Corresponding one-letter answers are provided. The questions are classified into five categories: 
\begin{itemize}
    \item \textbf{D} Replication of Definitions, $153$ question-answer pairs
    \item \textbf{F} Replication of Physical Facts, $138$ question-answer pairs
    \item \textbf{C} Conceptual Physics and Qualitative Reasoning, $238$ question-answer pairs
    \item \textbf{S} Single-Step Reasoning, $223$ question-answer pairs
    \item \textbf{M} Multi-Step Reasoning, $71$ question-answer pairs
\end{itemize}

We emphasize that this is the first dataset to our knowledge, which contains a large number of question and answer pairs tackling modern day physics topics up to the proficiency level of a physics bachelor degree or advanced high school degree. In this way, the dataset can help to push our understanding of how well LLMs are informed about the physical world in a language aligned with our current scientific description of it. Two datasets coming close to this in spirit are \citep{bisk2019piqareasoningphysicalcommonsense, wang-etal-2023-newton}. However they both focus on situative physics effects rather than a vast range of topics in physics as a science. With this, the dataset used here provides a unique opportunity to study the compressed physics knowledge of LLMs and (hypothetically) their reasoning capabilities thereof.

For examples of the dataset, we provide one question and answer pair for each question category in the appendix \ref{app:ssec:examples}.

\subsection{Models}
\label{ssec:models}

To facilitate our analysis, we accessed open-source models on the BlaBlador server infracture provided by FZ Jülich (Germany). The infrastructure stores and runs a variety of LLM models. We accessed these models through a REST API mechanism compliant with the \href{https://github.com/openai/openai-python}{\texttt{openai-python} library}. In this fashion, we were able to compare the performance of four LLMs: Llama3.1-8B-Instruct, Mixtral-8x7B-Instruct-v0.1, Mistral-7B-Instruct-v0.3, and GPT-3.5-turbo. We have set the temperature parameter for all models to a fixed value of 0.7.

All models were evaluated using a fixed few-shot prompting approach. They were asked one question at a time as the user input field. For each question or repetition of a question, the chat session was reset. To reduce the complexity of evaluation, we instructed the models to respond with only the letter corresponding to the correct answer.

We manually created examples for few-shot prompting similar to those in the dataset, see appendix \ref{app:sec:fsprompt}. To achieve this, we followed guidelines in \cite{llama_prompting_2024}. After testing zero-shot, one-shot, two-shot, and three-shot prompting, we chose the three-shot approach. This method proved effective as it allowed not only Llama but also other models to follow instructions accurately. To create examples, we requested GPT-4 to generate similar samples based on the real dataset. 
%if the shots in few-shot prompting were taken from the same dataset, our accuracy is overestimated. Let's keep that in mind.

Since different models are sensitive to various styles of prompting, we filtered out whitespace, newline characters, and any potential explanations using regular expressions. Additionally, in a few instances, the Mistral model returned two letters as a response. In such cases, or when the response did not match the expected pattern, the replies were replaced with \texttt{None} and excluded from further analysis.

\subsection{Uncertainty Estimation}

% In this work, we aim to evaluate the uncertainty of the LLM when providing answers to specific questions. To achieve this, we repeatedly prompt the model with the same question \( N = 20 \) times. This procedure is conducted for all questions in the \texttt{mlphys101} dataset. We measure the diversity among the responses $y$. We use the entropy of the answers \( Y \) given the question \( x \) and the prompt \( h \) as a metric for diversity and hence uncertainty: $H(Y|x,h) = -\sum_{y}p(y|x,h) \ast ln[p(y|x,h)]$.

% % \begin{equation}
% %     H(Y|x,h) = -\sum_{y}p(y|x,h) \ast ln[p(y|x,h)]
% %     \label{eqn:entropy}
% % \end{equation}

% As this is a discrete scenario of $5$ possible answers with \( N \) samples per question $x$, we estimate the probability of each answer $p$ by calculating the ratio of the number of occurrences of each letter for a particular question to the total number of times the question was asked: $ p (y = one\_letter\_answer|x,h) = \frac{count(one\_letter\_answer)}{N}$. These probabilities are then used as input to the entropy calculation stated above.%of equation \ref{eqn:entropy}.

In this work, we aim to evaluate the uncertainty of a LLM when generating answers to specific questions. To achieve this, we prompt the model with each question from the \texttt{mlphys101} dataset, repeating each prompt \( N = 20 \) times to gather N responses on the same question. For each question, we then assess the diversity in the model's answers by calculating the frequency with which each answer choice \( y_i \) (\( y_i \) $\in$ \(\{A, B, C, D, E\}\)) appears across the \( N = 20 \) responses. This frequency serves as an approximation of the probability for each response choice in our discrete setting: $ p (y_i|x,h) = \frac{count(y_i)}{N}$.

Using these probabilities, we compute the entropy \( H(Y|x, h) \) of the model's responses \( Y \) to quantify the uncertainty for each question \( x \) when prompted with a specific three-shot prompt \( h \):\[H(Y|x,h) = - \sum_{i} p(y_i|x,h) \ast ln[p(y_i|x,h)]
\]

This entropy measure allows us to gauge the level of consistency or uncertainty in the model’s answers to individual questions.

\section{Results and Discussion}

\begin{figure}[ht!]
  \centering
  \includegraphics[width=0.95\textwidth]{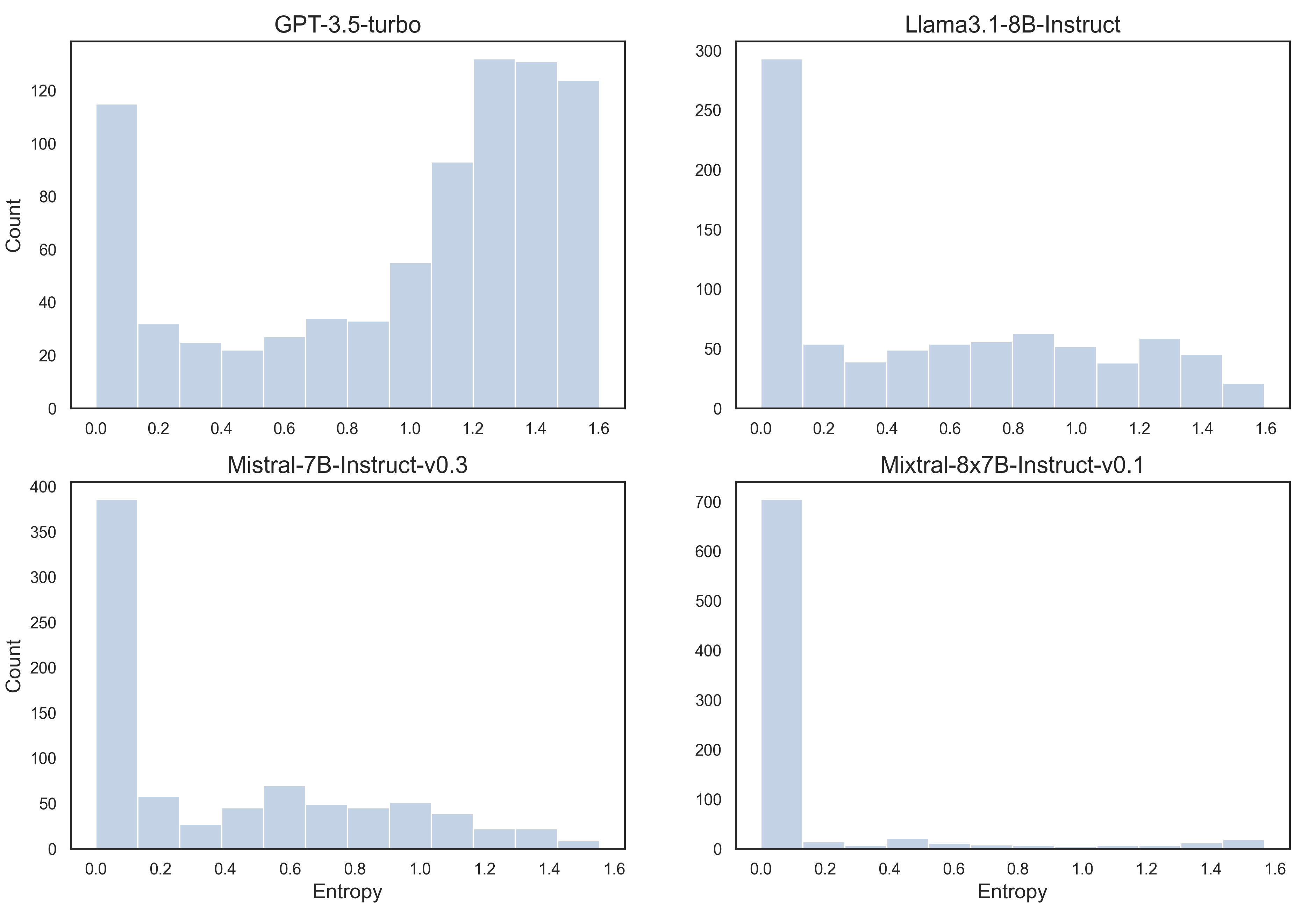}
  
  \caption{Entropy obtained from the distribution of answers to single questions of the \texttt{mlphys101} dataset \citep{mlphys101} for all four models.%Low entropy indicates a peaked distribution of answers, i.e. replies focus on a singular answer. High entropy indicates a broad distribution of answers for the same question.
  }
  \label{hist}
\end{figure}

In Figure \ref{hist}, we show an overview of diversity of answers in all models regardless of whether the answers are correct or incorrect. For Llama3.1-8B-Instruct, Mixtral-8x7B-Instruct-v0.1, and Mistral-7B-Instruct-v0.3, our analysis shows a concentration of replies at entropy close to $0$. Thus, these models appear more likely to provide responses with low entropy (low diversity, high consistency). In contrast, GPT-3.5-turbo tends to produce responses with higher diversity, i.e. more entries with entropy $H \geq 1$. In addition, Mixtral-8x7B-Instruct-v0.1 (being the largest model in use for our analysis) demonstrates a majority of entries in the lowest entropy bin and thus provides answers with minimal (if not vanishing) diversity. We can hypothesis at this point, that the degree of reliability is lowest with GPT-3.5-turbo as all question-answer pairs only exhibit one correct answer.

\begin{figure}[ht]
  \centering
  \includegraphics[width=\textwidth]{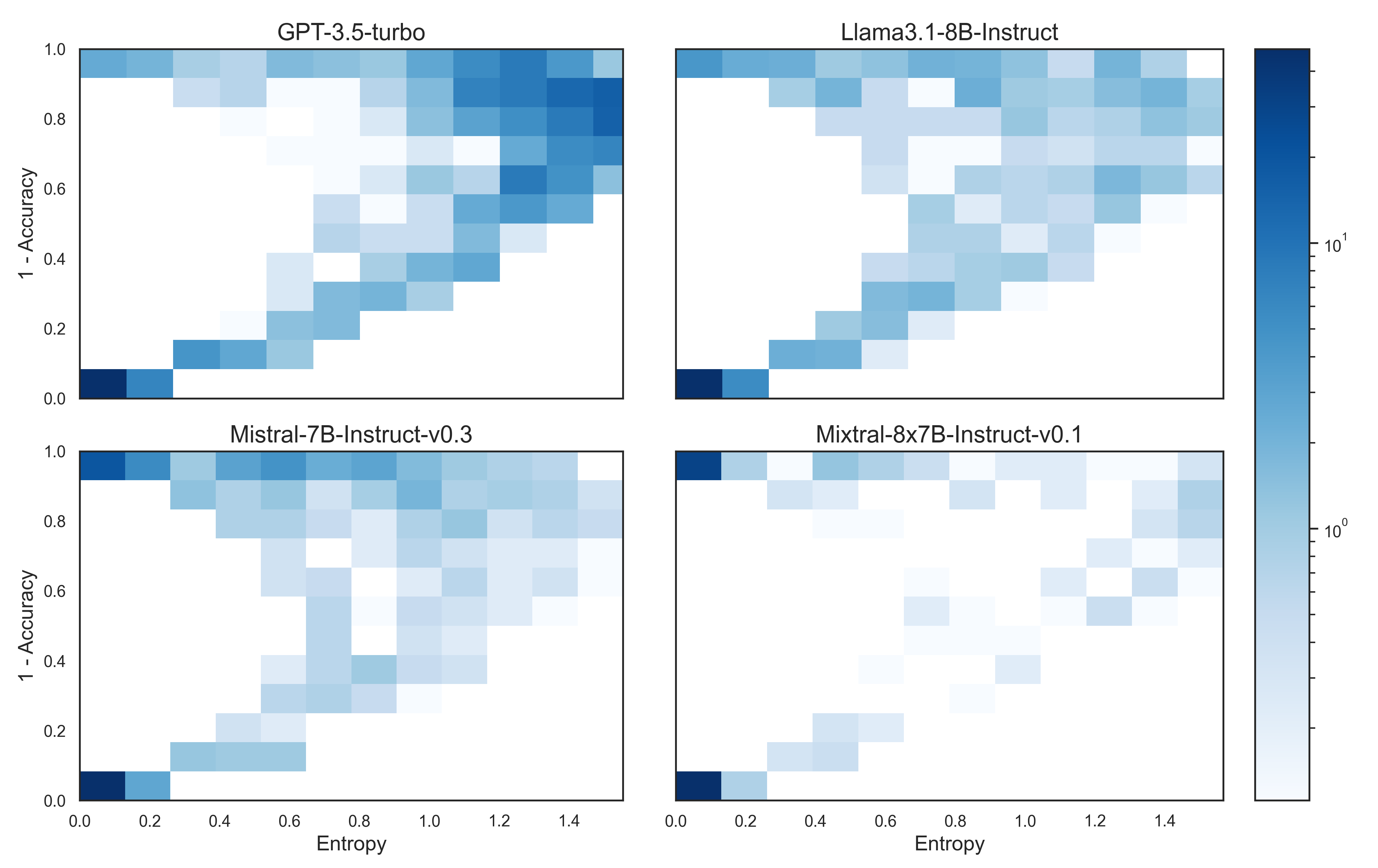}
  \caption{Two-dimensional Histogram of Error Rate (1 - Accuracy) vs. Entropy across Models. The binning of entropy is identical to figure \ref{hist}.}
  \label{curve}
\end{figure}

In order to study this hypothesis, we calculated and compared the error rate in responses versus the entropy of responses. This can provide direct insights in the degree of hallucination as LLM users assume the model to be correct no matter how often a question is raised. Figure \ref{curve} summarizes the results of this study.

The 2D histograms of ''1-Accuracy'' (or Error Rate) versus Entropy for all models exhibit a similar bell-shaped distribution (for a detailed mathematical explanation of the curve's shape, refer to Appendix \ref{app:sec:math}). In the bottom-left corner are questions where the models provide highly accurate answers with very low uncertainty. Conversely, in the top-right corner, questions are located where models give low accuracy responses with high uncertainty—indicating hallucinated answers. However, questions in the top-left corner reveal instances where models provide incorrect results with high certainty - which can also be attributed to the effect of hallucination. Figure \ref{curve} also illustrates that not all models are created equal in this regard. Diversity is lowest in the results of Mixtral-8x7B-Instruct-v0.1. Diversity is highest with GPT-3.5-turbo. We refer the curious reader to figure \ref{count} for a detailed account of this analysis.

Additionally, we examined the accuracy-certainty trade-off for each LLM across the five question categories (see Appendix \ref{app:sec:var-vs-cat}). Figure \ref{cat} in the appendix summarizes our findings. The overall shape of the accuracy-certainty curve (figure \ref{curve}) remains consistent across a majority of categories for all models below 10 billion parameters. A distinct behavior is visible, where diversity increases with the question category becoming more complex (complexity increases from \textbf{D} to \textbf{M}). For single-step \textbf{S} and multi-step reasoning questions \textbf{M}, GPT-3.5-turbo yields a minimal number of correct replies with high diversity. Mixtral-8x7B-Instruct-v0.1 can provide correct and incorrect answers at low diversity. The remaining models perform in the middle ground between these two extremes. We further suggest, that failure of LLMs in single and multi-step reasoning questions is inline with findings from other fields \citep{nezhurina2024alicewonderlandsimpletasks, lu2024emergentabilitieslargelanguage}.

One of the main limitations of this work stems from the prompting approach. Previous studies have shown that response accuracy is highly dependent on the prompting context and style used \cite{machlab2024llmincontextrecallprompt}. This might explain the high diversity observed in the entropy plot for the GPT-3.5-turbo model, suggesting the need for further experimentation with different prompting techniques. However, we hypothesize that the observed variability is unlikely to affect the overall shape of the error rate versus entropy curve. This consideration requires validation through additional testing and comparisons to other datasets.

\section{Summary}

% Our work demonstrates that uncertainty quantification is an essential step towards a trustworthy use of LLMs. We observe an accuracy-uncertainty trade-off in answers to a physics question-answer dataset for three of four models under study. This effect depends on model size and question difficulty.%With increasing difficulty, the LLM we used demonstrate vanishing usefulness for practical applications.  
% %\clearpage

Reliability of LLMs is an important component towards their trustworthiness. Hallucinations and inconsistency of model’s reply may lead to incorrect replies and losing users’ trust. Hallucinations may often appear in narrow domains possibly due to the lack of training data. For this reason we have proposed a pipeline for evaluating an accuracy-uncertainty trade-off. We tested it on  a physics MCQ dataset for four popular LLMs. The dataset exposes questions and answer pairs at different levels of complexity and reasoning demand. The experiment has shown the difference in consistency of responses and hallucinating depending on model size and question complexity. A downstream analysis has to be undertaken, to identify the root cause of these observations and compare these findings to different datasets of similar nature.

\clearpage

%\section*{References}
\bibliography{references}

%%%%%%%%%%%%%%%%%%%%%%%%%%%%%%%%%%%%%%%%%%%%%%%%%%%%%%%%%%%%
\appendix

\section{Appendix}
\label{app}

\subsection{Dataset Examples}
\label{app:ssec:examples}

As discussed in section \ref{ssec:dataset}, our dataset consists of $5$ classes of questions. We provide one example for each here to provide a more comprehensive insight into the variety of topics and requirements for an answer.

\subsubsection{Replication of Definitions, \textbf{D}}

\begin{description}
    \item[Question] The property of a moving object to continue moving is what Galileo called 
    \item[Answer A] velocity.
    \item[Answer B] speed. 
    \item[Answer C] acceleration. 
    \item[Answer D] inertia. 
    \item[Answer E] direction.
\end{description}

\subsubsection{Replication of Physical Facts, \textbf{F}}

\begin{description}
    \item[Question] \_\_\_\_ are examples of vector quantities.. 
    \item[Answer A] Acceleration and time
    \item[Answer B] Velocity and acceleration 
    \item[Answer C] Volume and velocity 
    \item[Answer D] Mass and volume 
    \item[Answer E] Time and mass
\end{description}

\subsubsection{Conceptual Physics and Qualitative Reasoning, \textbf{C}}

\begin{description}
    \item[Question] If an object is moving, then the magnitude of its \_\_\_\_ cannot be zero.
    \item[Answer A] speed
    \item[Answer B] velocity 
    \item[Answer C] acceleration 
    \item[Answer D] A and B 
    \item[Answer E] A, B, and C
\end{description}

\subsubsection{Single-Step Reasoning, \textbf{S}}

\begin{description}
    \item[Question] A firefighter with a mass of 70 kg slides down a vertical pole, accelerating at $2 m/s^2$. The force of friction that acts on the firefighter is
    \item[Answer A] $70\,N$. 
    \item[Answer B] $560\,N$. 
    \item[Answer C] $140\,N$. 
    \item[Answer D] $700\,N$. 
    \item[Answer E] $0\,N$.
\end{description}

\subsubsection{Multi-Step Reasoning, \textbf{M}}

\begin{description}
    \item[Question] A bowling ball at a height of 36 meters above the ground is falling vertically at a rate of 12 meters per second. Which of these best describes its fate?
    \item[Answer A] It will hit the ground in exactly three seconds at a speed of $12 m/s$.
    \item[Answer B] It will hit the ground in less than three seconds at a speed greater than $12 m/s$.
    \item[Answer C] It will hit the ground in more than three seconds at a speed less than $12 m/s$. 
    \item[Answer D] It will hit the ground in less than three seconds at a speed less than $12 m/s$. 
    \item[Answer E] It will hit the ground in more than three seconds at a speed greater than $12 m/s$.
\end{description}

\section{Few-shot prompting}
\label{app:sec:fsprompt}

We use few-shot prompting to trigger the LLM for an answer. Here is an example prompt to illustrate our strategy.

\begin{lstlisting}
system(''You're a highly knowledgeable physics tutor. For each message, 
give only the letter of the correct answer without any
explanations or additional information.''),

user(''A ball rolls down a slope and accelerates uniformly 
at 2 m/s^2. If it starts from rest, what will be its speed after
3 seconds? A. 3 m/s, B. 4 m/s, C. 5 m/s, D. 6 m/s, E. 7 m/s''),

assistant(''D''),

user(''A cyclist accelerates uniformly from rest to a speed 
of 10 m/s in 5 seconds. What is their acceleration? 
A. 1 m/s^2, B. 2 m/s^2, C. 3 m/s^2, D. 4 m/s^2, E. 5 m/s^2''),

assistant(''B''),

user(''A rocket accelerates from rest at a constant rate of 
6 m/s^2. What speed will it reach after 4 seconds? 
A. 12 m/s, B. 18 m/s, C. 24 m/s, D. 30 m/s, E. 36 m/s''),

assistant(''C''),

user(''<question to evaluate goes here>''),

\end{lstlisting}

Details on how the few shot prompts were created are given in section \ref{ssec:models}.

\section{Variability of Answers given Question categories}
\label{app:sec:var-vs-cat}

\begin{figure}[ht!]
  \centering
  \includegraphics[width=\textwidth]{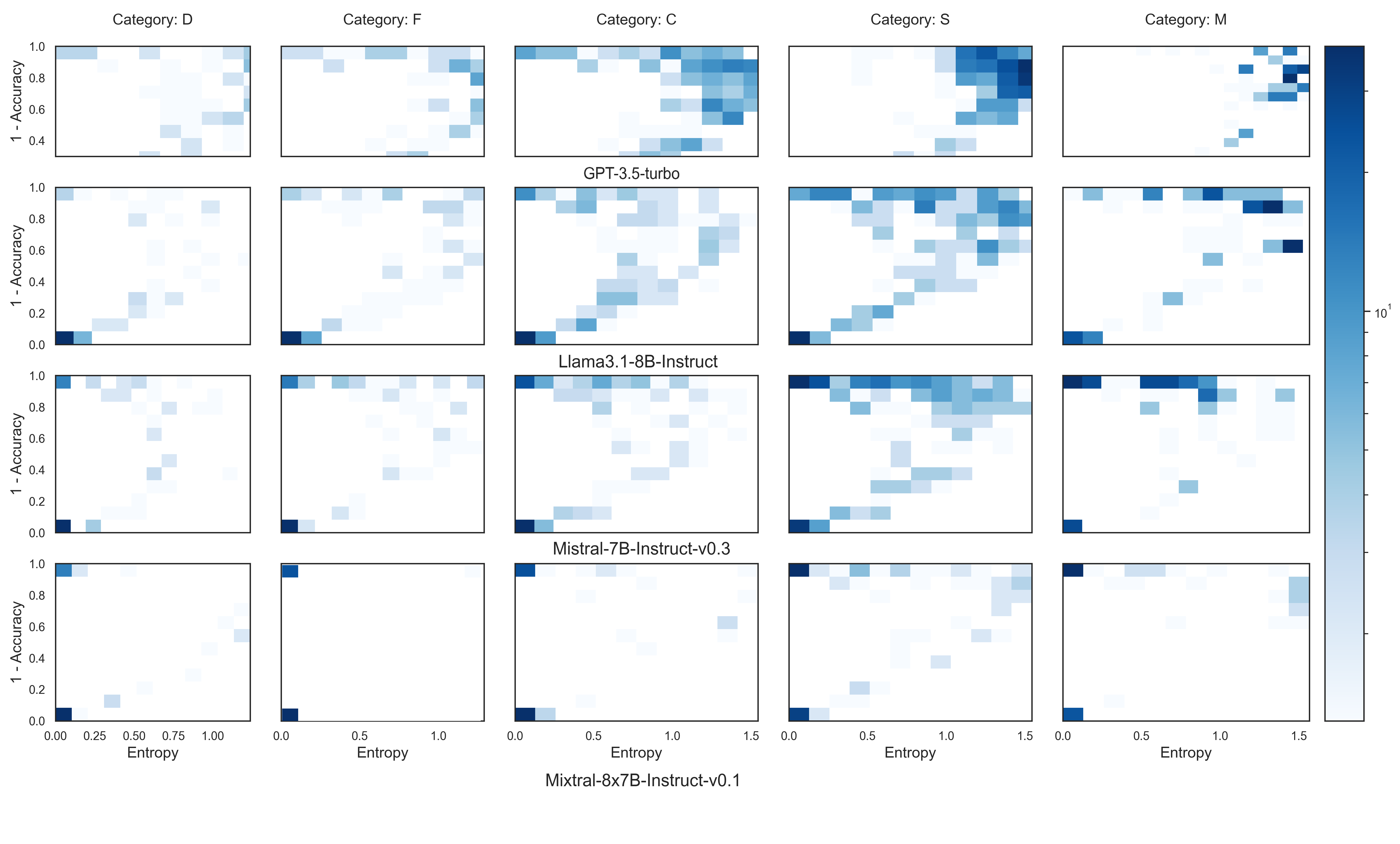}
  \caption{Accuracy-certainty trade-off for each LLM in five question categories}
  \label{cat}
\end{figure}

\begin{figure}[ht!]
  \centering
  \includegraphics[width=\textwidth]{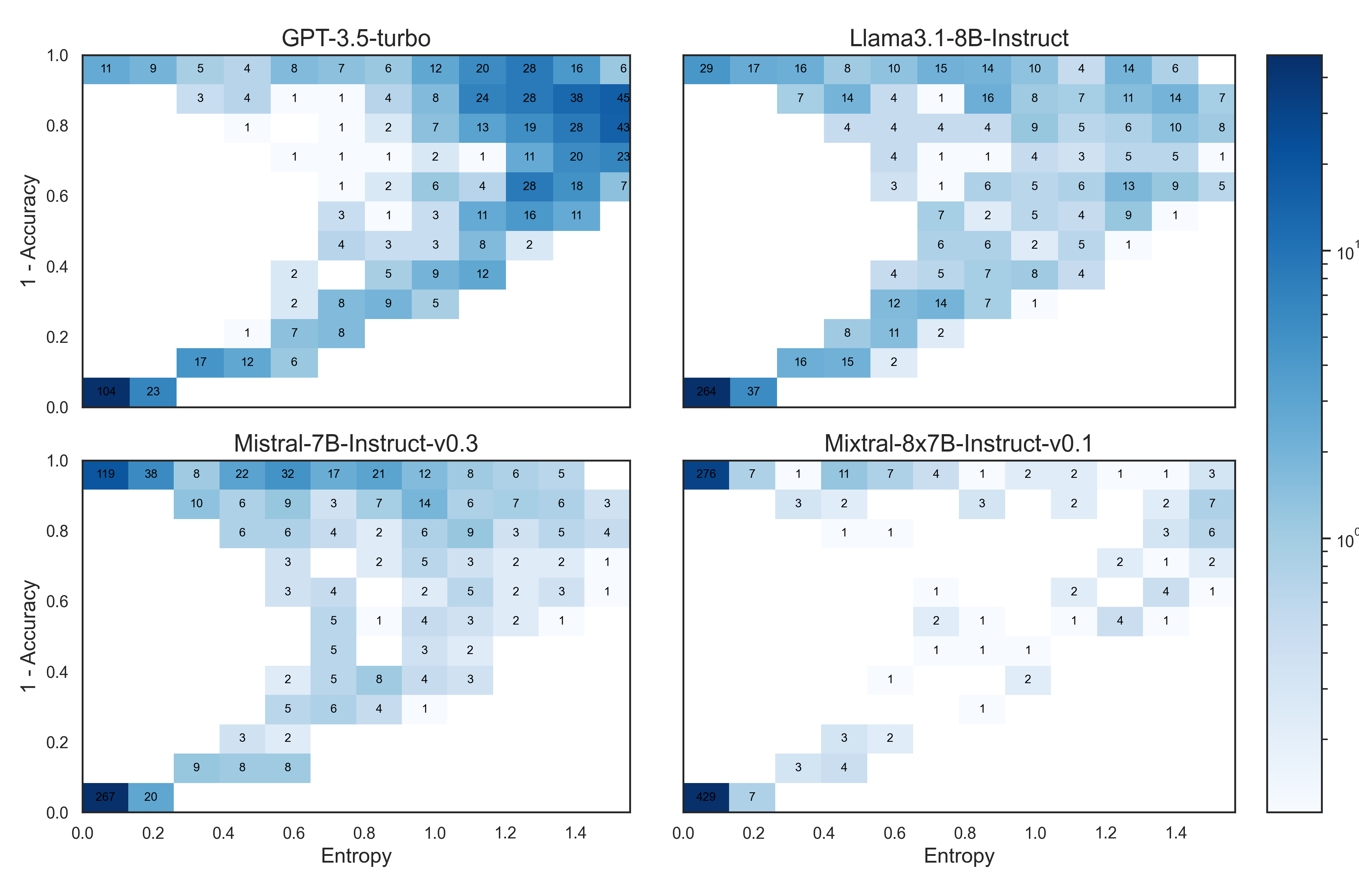}
  \caption{Two-dimensional Histogram of Error Rate (1 - Accuracy) vs. Entropy across Models with counts per bin. Entries are identical to figure \ref{curve}.
}
  \label{count}
\end{figure}

\section{Code and Data Availability}
\label{app:sec:code}

We make our analysis software available at \href{https://github.com/ReganovaLisa/Testing_Large_Language_Models_for_Physics_Knowledge}{here}. The \texttt{mlphys101} dataset is available \href{https://rodare.hzdr.de/record/3137}{here} until it has been published in a peer-reviewed journal.

\section{Shape of the Curves}
\label{app:sec:math}
To analyze the shape of the curve, let us consider different scenarios of a model's response patterns separately:

\begin{enumerate}
    \item The model generates only correct responses for a given question.
    \item The model generates only incorrect responses.
    \item The model generates two distinct responses, one of which is correct and the other incorrect.
    \item The model generates three or more distinct responses, with at least one being correct.
\end{enumerate}

When the model consistently produces correct responses, both the entropy and the error rate are zero (accuracy equals 1). On the curve, this scenario corresponds to the bottom-left corner. A higher density of points in this region indicates that the model often generates accurate and consistent responses.

When the model fails to generate any correct responses, all points align along the horizontal line where the error rate equals 1. The top-left corner of the curve represents scenarios where the model's responses are consistently incorrect.

In the case where the model produces only two different responses - one correct and one incorrect — the entropy $H(Y | x, h)$ is given by:
\[
H(Y | x, h) = - p(y_{\text{correct}} | x, h) *\ln[p(y_{\text{correct}} | x, h)] - p(y_{\text{incorrect}} | x, h) *\ln[p(y_{\text{incorrect}} | x, h)].
\]

Substituting $p(y_{\text{correct}} | x, h) = \text{accuracy}$ and $p(y_{\text{incorrect}} | x, h) = 1 - \text{accuracy}$, the entropy can be expressed as:
\[
H(Y | x, h) = -  (\text{accuracy}) *\ln(\text{accuracy}) - (1 - \text{accuracy}) * \ln(1 - \text{accuracy}), \]
\begin{equation} \label{eq1}
H(Y | x, h) =  - (1 - \text{error\_rate}) * \ln(1 - \text{error\_rate}) - \text{error\_rate} * \ln(\text{error\_rate}). 
\end{equation}

\begin{figure}[ht!]
  \centering
  \includegraphics[width=0.95\textwidth]{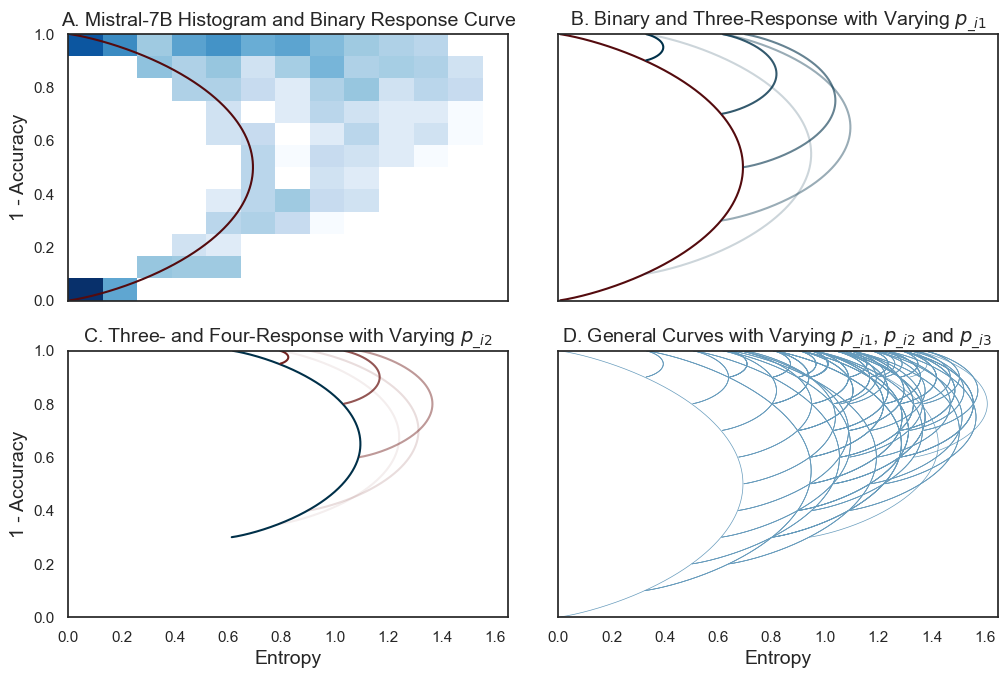}
  \caption{A. Two-dimensional histogram of (1 - Accuracy) vs. Entropy for the Mistral 7B model, shown alongside the theoretical curve (red) representing the scenario where the model provides only two distinct responses, one of which is correct (see Equation\ref{eq1}).
B. Theoretical curves for binary responses (red) and three distinct responses (blue) with $p_{\_i1}$ = \{0.1, 0.3, 0.5, 0.7, 0.9\} (see Equation \ref{eq2}). Curve intensity increases as $p_{\_i1}$ increases.
C. Theoretical curves (Equation \ref{eq3}) for three (blue) and four (red) distinct responses, with $p_{\_i1}$ = 0.3 and $p_{\_i2}$ = \{0.05, 0.1, 0.3, 0.5, 0.65\}. Curve intensity increases as $p_{\_i2}$ increases.
D. Theoretical curves based on the general equation  \ref{eq:4} with parameters $p_{\_i1}$, $p_{i2}$, and $p_{i3} $ varying within \{0, 0.1, 0.2, 0.3, 0.4, 0.5, 0.6, 0.7, 0.8, 0.9, 1.0\}.
}
  \label{math}
\end{figure}

The resulting curve is illustrated in Figure \ref{math}(A). This pattern can also be observed in the curves for the Mistral and Llama models (see Figure \ref{hist}).

When a model's responses include correct replies and more than one distinct incorrect replies (i.e., two or more versions of incorrect output), we obtain families of parameterized curves with \((\# \text{ of incorrect types} - 2)\) parameters. 

For example, in the simplest case, let us consider three distinct responses:
\(p(y_{\text{correct}} | x, h)\), \(p(y_{\text{incorrect\_1}} | x, h)\), and \(p(y_{\text{incorrect\_2}} | x, h)\).  
The probabilities can be defined as follows:
\[
p(y_{\text{correct}} | x, h) = \text{accuracy} = 1 - \text{error\_rate},\] \[ 
p(y_{\text{incorrect\_1}} | x, h) = p(y_{\text{incorrect\_1}} | x, h) = p_{\text{\_i1}},  \] \[ p(y_{\text{incorrect\_2}} | x, h) = \text{error\_rate} - p_{\text{\_i1}}.\]

In this case, the entropy \(H(Y | x, h)\) is given by:
\[
H(Y | x, h) = - (1 - \text{error\_rate}) * \ln(1 - \text{error\_rate}) 
- p_{\text{\_i1}} * \ln(p_{\text{\_i1}}) - \]
\begin{equation} \label{eq2}
- (\text{error\_rate} - p_{\text{\_i1}}) * \ln(\text{error\_rate} - p_{\text{\_i1}}).
\end{equation}

Examples of such curves, for \(p_{\text{\_i1}} = \{0.1, 0.3, 0.5, 0.7, 0.9\}\), are shown in Figure \ref{math}(B).

Similarly, the equations for four and five distinct responses among replies to a single query can be expressed as follows:  

For four types of responses:
\[
H(Y | x, h) = -(1 - \text{error\_rate}) * \ln(1 - \text{error\_rate}) 
- p_{\text{\_i1}} * \ln(p_{\text{\_i1}}) - \]
\begin{equation} \label{eq3}
-  p_{\text{\_i2}} * \ln(p_{\text{\_i2}}) 
- (\text{error\_rate} - p_{\text{\_i1}}
-  p_{\text{\_i2}}) * \ln(\text{error\_rate} - p_{\text{\_i1}} - p_{\text{\_i2}}).
\end{equation}

For five types of responses:
\[
H(Y | x, h) = -(1 - \text{error\_rate}) * \ln(1 - \text{error\_rate}) 
- p_{\text{\_i1}} * \ln(p_{\text{\_i1}}) - p_{\text{\_i2}} * \ln(p_{\text{\_i2}}) 
- p_{\text{\_i3}} * \ln(p_{\text{\_i3}}) - \]
\begin{equation} \label{eq:4}
- (\text{error\_rate} - p_{\text{\_i1}} - p_{\text{\_i2}} - p_{\text{\_i3}}) * 
\ln(\text{error\_rate} - 
- p_{\text{\_i1}} - p_{\text{\_i2}} - p_{\text{\_i3}}).
\end{equation}

All the equations previously described in this section are special cases of Equation \ref{eq:4}, where some of the probabilities are equal to zero (see Figure \ref{math}(D)).

%%%%%%%%%%%%%%%%%%%%%%%%%%%%%%%%%%%%%%%%%%%%%%%%%%%%%%%%%%%%

\end{document}